\documentclass{article}

\usepackage{arxiv}

\usepackage{comment}
\usepackage[T1]{fontenc}    
\usepackage[hidelinks]{hyperref} 
\usepackage{url}            
\usepackage{booktabs}       
\usepackage{graphicx}
\graphicspath{{graphics/}}
\usepackage{tabularx}
\usepackage{authblk}

\title{Review of multimodal machine learning\\ approaches in healthcare}

\author[1]{Felix Krones}
\author[2]{Umar Marikkar}
\author[1]{Guy Parsons}
\author[3]{Adam Szmul}
\author[1]{Adam Mahdi}
\affil[1]{Oxford Internet Institute, University of Oxford, Oxford, UK}
\affil[2]{Institute for People-Centred AI, University of Surrey, Guildford, UK}
\affil[3]{Wellcome/EPSRC Centre for Interventional and Surgical Sciences, 
University College London, London, UK}

\renewcommand{\shorttitle}{Review of multimodal machine learning approaches in healthcare}

\hypersetup{
    pdftitle={Review of multimodal machine learning approaches in healthcare},
    pdfauthor={Felix H. Krones, Umar Marikkar, Dr Adam Szmul, Guy Parsons, Dr Adam Mahdi},
    pdfkeywords={data fusion, medical imaging, self-supervised learning, multimodal machine learning},
}

\date{}

\begin{document}
\maketitle

\begin{abstract}
    Machine learning methods in healthcare have traditionally focused on using data from a single modality, limiting their ability to effectively replicate the clinical practice of integrating multiple sources of  information for improved decision making. Clinicians typically rely on a variety of data sources including patients' demographic information, laboratory data, vital signs and various imaging data modalities to make informed decisions and contextualise their findings. Recent advances in machine learning have facilitated the more efficient incorporation of multimodal data, resulting in applications that better represent the clinician's approach. Here, we provide a review of multimodal machine learning approaches in healthcare, offering a comprehensive overview of recent literature. We discuss the various data modalities used in clinical diagnosis, with a particular emphasis on imaging data. We evaluate fusion techniques, explore existing multimodal datasets and examine common training strategies.

\end{abstract}


\section{Introduction}
Artificial Intelligence (AI), and machine learning in particular, has radically altered our interactions with the world, fostering rapid advancements in various domains. However, the adoption of machine learning approaches in healthcare has been slower than in other fields despite the increasing pressure on healthcare systems and the urgent demand for high quality, personalised care \cite{kirch2017addressing, topol2019high}. 

Machine learning methods can already outperform humans in some areas, especially when time restrictions apply \cite{topol2019high,bejnordi2017diagnostic} and their integration into clinical workflows presents a significant opportunity to improve healthcare and alleviate challenging resource constraints \cite{bartoletti2019ai}.

Advanced machine learning techniques have supported the integration of multimodal healthcare data, allowing the creation of methods that mimic clinical practice by incorporating diverse information sources into decision-making processes. Clinicians often consider multiple data sources when assessing a patient's condition and determining appropriate treatment, taking advantage of the additional information and context this approach supports. For example, a study of radiologists found that 85\% deemed the clinical context to be crucial for the interpretation of radiological examinations \cite{Zhou2021}. Patients' demographic data, medical history, laboratory data and vital signs can all be used by physicians to contextualise their assessments of medical images within the broader clinical picture. In response to this, an increasing volume of AI research in healthcare has focused on the use of multimodal data, aiming to better emulate clinicians' approaches and enhance overall performance \cite{soenksen2022integrated}.

Many studies have explored the application of machine learning to healthcare and medicine, investigating in particular the development of deep learning techniques across various modalities. These investigations have encompassed a broad range of topics, including general applications \cite{Piccialli2021, shamshirband2021review, Esteva2019, Akay2019, Zhang2018, Fatima2017}, those placing particular emphasis on image-based approaches  \cite{Shen2017, Litjens2017, wang2021review} and a specific focus on the chest region  \cite{Bizopoulos2018, Krittanawong2019, An2018, ccalli2021deep}.
In the context of multimodal applications, some review papers have discussed application-agnostic approaches \cite{Baltrusaitis2018}, while others have emphasised fusion in disease diagnosis \cite{cui2022deep}. Numerous reviews have delved into models and architectures, addressing their optimisation strategies in general \cite{Gao2020, Ramachandram2017} and specifically within the healthcare domain \cite{behrad2022overview, Xu2021, huang2020fusion, Huang2020_2, Heiliger2022, Stahlschmidt2022, Ayesha2021}. Several reviews have also explored the potential applications of multimodal machine learning, often highlighting the inclusion of image modalities within the multimodal pipeline \cite{acosta2022multimodal, lipkova2022artificial, amal2022use, kline2022multimodal}. 
Other reviews have centred on self-supervised learning, following developments in that area \cite{wojcik2022foundation, krishnan2022self, fei2022towards, shurrab2022self}.

Here we provide a comprehensive examination of multimodal machine learning data fusion techniques in healthcare. Our review explores the primary data modalities often used in data fusion, highlighting the essential steps involved in the development of multimodal machine learning models. We offer an up-to-date, detailed overview of commonly used multimodal datasets and an extensive survey of studies using multimodal approaches across a wide range of medical conditions.

\section{Data modalities}
\begin{figure}[t!]
    \centering
    \includegraphics[width=1.0\textwidth]{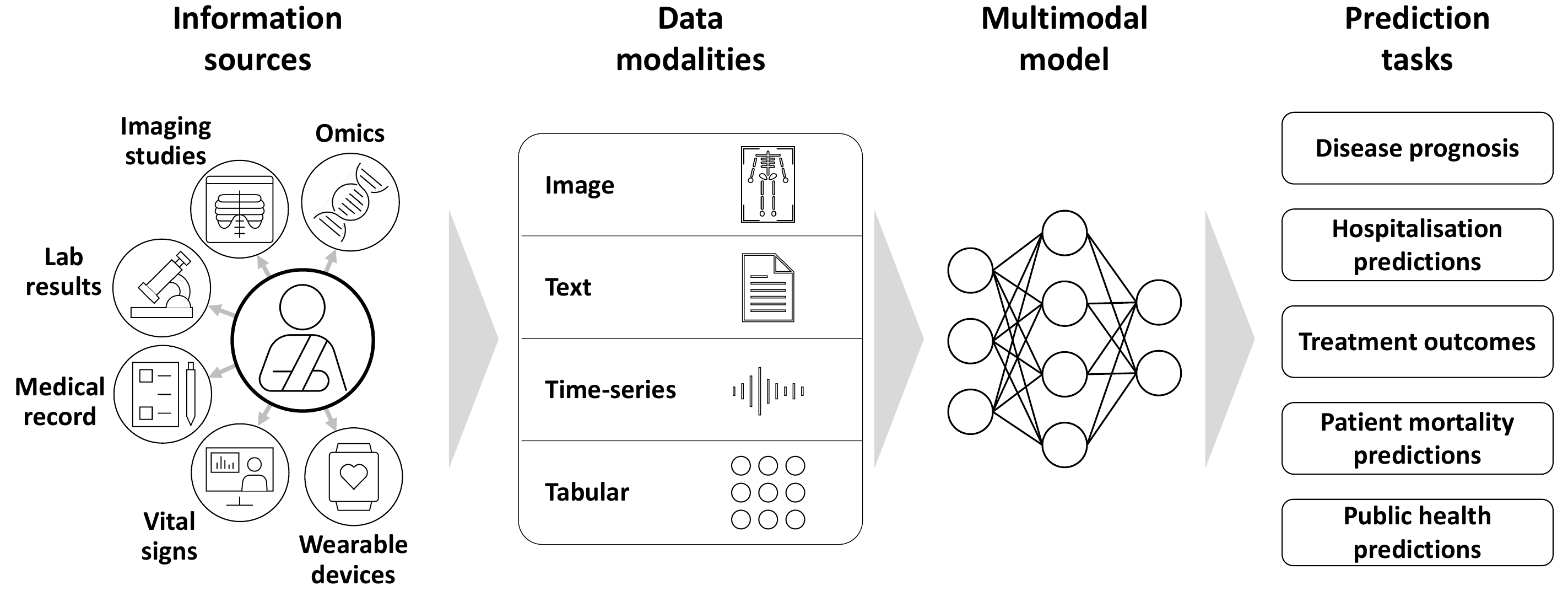}
    \caption{\textbf{Clinical data modalities and prediction tasks.} Distinct data modalities play pivotal roles in clinical decision-making: imaging data, text data, time-series data and tabular data. All are used for various clinical predictions tasks. Medical prediction tasks in clinical practice involve leveraging machine learning models and algorithms to forecast future clinical outcomes based on existing patient data. They play a crucial role in the decision-making process for diagnosis, prognosis and treatment.}
    \label{fig:medical_modalities}
\end{figure}

Multiple data modalities find application in clinical practice \cite{qayyum2020secure, duvieusart2022multimodal} as illustrated in \autoref{fig:medical_modalities}. These include imaging data (e.g. X-rays), text data (e.g. radiology reports), time-series data (e.g. vital signs) and cross-sectional or panel tabular data (e.g. metadata). Recognising the critical role of data selection in multimodal machine learning, we provide an overview of the key data modalities used in modeling studies.

\subsection{Imaging data}
Most medical imaging data are stored as 2D image slices in the Digital Imaging and Communications in Medicine (DICOM) format after acquisition  \cite{bidgood1997understanding}. This includes patients' metadata, details regarding the imaging procedure, information about the device used for image acquisition and imaging protocol settings. Medical 3D volumetric images are typically constructed from a stack of 2D slices with a specified thickness, representing a specific region of interest within the body. These individual slices can be processed and analysed separately (as 2D images) or collectively (as 3D volumes) to extract valuable information.

When preparing medical imaging data for machine learning, data in the DICOM format are either converted to commonly used image formats, such as JPEG or PNG \cite{tan2006image}, or to the Neuroimaging Informatics Technology Initiative (NifTi) format \cite{poldrack2019computational}, a dedicated medical image analysis format where, along with the image, only essential metadata are stored in the header.

\smallskip

\textbf{X-ray imaging.} 
Medical X-ray imaging is a widely available and inexpensive 2D imaging technique. Almost half of the 43.3 million imaging scans in England in 2022 were X-ray images, making it the most commonly used medical imaging modality \cite{england2016diagnostic}. X-ray imaging operates on the principle of differential attenuation of X-rays through various types of body tissues \cite{dendy2011physics}. X-rays are inherently 2D grey-scale radiograph images. Conventional radiographs identify five levels of attenuation: air, fat, soft tissue, bone and metal. In radiographs, air appears the darkest because its low density allows most X-rays to pass through, whereas metal, being significantly denser, appears bright white due to its absorption of the majority of the X-ray beam's energy. Fat tissue, soft tissue and bone manifest as varying shades of grey, with fat being darker than soft tissues and bone lighter \cite{Brant2012}. 

In machine learning applications, images are resized to specific fixed dimensions, such as $224\times224$ (e.g. the original VGGNet was trained with a resolution of 224 pixels \cite{simonyan2014very}), $256\times256$ or $512\times512$ pixels. Subsequently, depending on the model, grey-scale X-ray images are either maintained as single-channel images or transformed into three-channel images by replicating the existing channel \cite{duvieusart2022multimodal}.

While bone imaging remains the most common application for X-ray scans, their utility extends to various other areas of the body, encompassing radiography, fluoroscopy and mammography (primarily used for breast cancer screening) \cite{Martensen2013, gotzsche2013screening}. Since X-rays lack sufficient spatial depth information for unequivocal diagnosis, they are frequently used as a screening tool. Many recent studies have used X-rays in cardiology prediction tasks, such as cardiomegaly diagnosis \cite{duvieusart2022multimodal, palepu2022tier}, pulmonary oedema assessment \cite{chauhan2020joint} or accessory conduction pathway analysis \cite{nishimori2021accessory}.

\smallskip

\textbf{Computed Tomography (CT).} 
Computed Tomography (CT) scans provide detailed cross-sectional images of a human body \cite{buzug2011computed}. These scans generate 3D image volumes by reconstructing multiple subsequent 2D slices from radiographic projections acquired from multiple angles. CT scans are depicted in Hounsfield Units (HU), which directly correspond to signal attenuation caused by tissue density with respect to water \cite{goldman2007principles}. 

As a popular choice for medical diagnostics, CT scans offer high-resolution imaging, wide availability, cost-effectiveness and speed. However, they expose patients to ionising radiation and have limitations in distinguishing soft tissues \cite{goldman2007principles}.

CT is a versatile imaging technique primarily used to identify structural abnormalities, detect tumors, diagnose heart diseases and image the head for various neurological conditions \cite{Adam2014, Schoepf2007, Bremner2005}. It is often used in respiratory \cite{dougan2015role} and cardiological studies \cite{xu2020accurately, fang2021deep, Samak2020}, in cancer diagnosis \cite{wiener2013test} and treatment planning \cite{battista1980computed, brunelli2009ers}.

\smallskip

\textbf{Magnetic Resonance Imaging (MRI)}. 
Unlike the previously discussed imaging techniques, Magnetic Resonance Imaging (MRI) is a non-ionising method \cite{Hashemi2010}. 
The patient is placed in a strong magnetic field, which aligns the protons' magnetic moments (spin) of their body with the field.  
Short radio-frequency pulses disturb the alignment of these protons and then realign with the magnetic field. MRI measures magnetisation in both longitudinal and transverse directions, enabling tissue-specific reconstructions \cite{grover2015magnetic}. 

MRI maintains a high signal-to-noise ratio and provides a detailed view in two directions \cite{stadler2007artifacts}. However, it may lead to aliasing artifacts in the orthogonal direction, especially in 2D magnetic resonance acquisition protocols. To address such artifacts, magnetic resonance data may require anti-aliasing pre-processing steps, such as filtering or the application of dedicated machine learning models \cite{stadler2007artifacts}.

This technique works well for providing detailed visuals of soft tissues organs and internal structures without employing ionising radiation \cite{frisoni2010clinical, guermazi2013mri}. It is extensively used to study brain disorders, including Alzheimer's disease \cite{parisot2018disease, polsterl2021combining}, multiple sclerosis \cite{Yoo2019} and Parkinson's disease \cite{ryman2020mri}.

\textbf{Nuclear medicine imaging techniques.} 
Positron Emission Tomography (PET) and Single Photon Emission Computed Tomography (SPECT) are nuclear medicine imaging techniques that detect gamma photons emitted from radioactive tracers in the body, offering insights into metabolic activity \cite{israel2019two} and blood flow/function \cite{mullani2008tumor}. 

In practice, these techniques generate dozens to hundreds of 2D slices acquired per scan \cite{Dwivedi2022}. The slices are evenly spaced, ensuring consistent distance within the scan. The intensities within the images have a relative value and each scan is often accompanied by a paired attenuation correction CT image. The main limitations of these imaging modalities include long acquisition times compared to CT and usually lower resolution than CT or MRI. 

Nuclear medicine imaging techniques are important for diagnosing diseases, monitoring treatments and studying health at the molecular level \cite{braman2021deep, Dwivedi2022}. PET can be useful in cancer diagnosis and for neurological conditions \cite{duclos2021pet} while SPECT can improve staging, prognosis determination and treatment planning \cite{israel2019two}. These techniques are often paired with CT and MRI, providing complementary insights in tasks such as Alzheimer's prediction with MRI \cite{el2020multimodal, Suk2014}, lung cancer prediction with CT \cite{Hyun2019} or tumor segmentation tasks using both modalities \cite{Guo2019}.

\textbf{Ultrasound.} 
Ultrasound imaging uses acoustic energy with frequencies above 20 kHz to visualise body structures \cite{Woo2002}. The data typically consist of a series of 2D frames \cite{arovac2011ApplicationOU}; however recent advancements now enable 3D and 4D (real-time 3D) imaging \cite{merz20123d}. Image resolution varies depends on transducer frequency; higher frequencies provide greater resolution but less depth penetration and so are more useful for superficial structures while lower frequencies provide greater depth at the cost of resolution. Image intensities correspond to tissue echogenicity (i.e. the ability to reflect sound waves), with different shades of grey representing different tissue densities \cite{Woo2002}. Doppler techniques are often incorporated to visualise blood flow and assess velocity, providing helpful colour overlays on the grey-scale images \cite{arovac2011ApplicationOU}.

Prior to applying machine learning techniques, ultrasound data are pre-selected, as the spatial position or orientation is not immediately apparent and relies on the operator's approach \cite{brattain2018machine}. This often involves frame selection and identifying the region of interest (ROI) which can be achieved through either manual or automated methods. Ultrasound data inherently exhibit speckling, a characteristic that may require compensation during pre-processing \cite{karaouglu2022removal}. 

Ultrasound's non-invasiveness and lack of ionising radiation make it a favoured choice for obstetrics and gynecology \cite{merz20123d}. It can be used to evaluate heart function, monitor blood flow \cite{brattain2018machine} and review many organs (e.g. liver, kidneys) for potential issues \cite{arovac2011ApplicationOU}. Beyond these applications, ultrasound plays an important role in tracking disease progression and guiding precise surgical procedures \cite{arovac2011ApplicationOU}.

\textbf{Dermoscopic images.} 
Dermoscopy is a non-invasive imaging technique that captures high-resolution images of the skin surface and is primarily used for the early detection of skin cancers such as melanoma \cite{vestergaard2008dermoscopy}. By employing polarised or non-polarised light, dermoscopy can visualise skin structures that are otherwise invisible to the naked eye. 

Dermoscopic images can be analysed using computer-aided diagnostic tools which leverage machine learning approaches to assist dermatologists in making more accurate and efficient diagnoses. Before presenting images to a machine learning model, regions of interest are usually segmented to remove redundant information \cite{Kawahara2019}. The images also need to be cleaned of unwanted artefacts, such as hair, gel bubbles and ink marks \cite{iqbal2021automated}. 
As they are photographic images, they usually have three channels (RGB) like natural images, compared to most medical imaging which has only one channel for intensity. 

Dermoscopic images are primarily employed in dermatology to visualise subsurface skin structures in the epidermis and dermis, aiding in the diagnosis and monitoring of skin lesions and tumors \cite{Yap2018,  Gessert2020}. The technique has proven valuable for distinguishing between benign and malignant lesions, improving diagnostic accuracy and reducing the need for invasive procedures like biopsies \cite{kittler2002diagnostic}.

\subsection{Text data}
The text modality constitutes a fundamental and extensively used form of medical information in healthcare \cite{spasic2020clinical, mustafa2021automated}. Various patient-specific text modalities exist, many of which are common in clinical practice. These include procedure notes, comprehensive clinical records generated by healthcare providers, that include progress reports and consultation notes. Prescription notes comprise detailed instructions for medication regimens, specifying drug names, dosages and usage guidelines \cite{li2015end}. Medical discharge notes, critical in patient care transitions, document a patient's hospitalisation experience, including vital information such as diagnoses, treatments and post-discharge plans \cite{johnson2023mimic}. Referral letters serve as communication bridges between healthcare providers, containing pertinent patient details and the rationale behind referrals. Radiology notes offer descriptions of findings extracted from diagnostic imaging studies, encompassing X-rays, CT scans and MRIs \cite{Huang2021, casey2021systematic, duvieusart2022multimodal}. Beyond patient-specific data, the text modality also encompasses general medical knowledge drawn from diverse sources, including medical journals, literature, medical websites and pharmaceutical labels \cite{xie2021survey}. 

Natural Language Processing (NLP) techniques have been used to extract relevant information from medical records \cite{sheikhalishahi2019natural, locke2021natural}. Once this information is extracted it can be converted into a structured format that can then be used in machine learning models. For traditional NLP methods, one of the key challenges is the need for subject matter experts to label features of interest in text sources, a time-consuming task. Various approaches have been used to address this bottleneck, such as active learning techniques to prioritise text for labelling \cite{chen2015study}. Transfer learning, data augmentation techniques and the use of synthetic clinical notes  offer further potential in this area \cite{walonoski2018synthea}. 

\subsection{Time-series data}
Time-series data refers to data points that are collected over consistent intervals of time. One common example is the Electrocardiogram (ECG), which traces the electrical activity of the heart over a short period of time, providing insights into heart rhythm and potential abnormalities \cite{ahsan2022machine}. 
Electroencephalography (EEG) records brain activity, enabling clinicians to detect neurological disorders and monitor treatment responses \cite{zheng2021predicting}. 
Foetal monitoring allows for the surveillance of a foetus's well-being during pregnancy and labour, highlighting any distress or deviations through the consistent recording of heart rate and other variables \cite{freeman2012fetal}. 
Intracranial pressure monitoring in patients with conditions such as traumatic brain injury can help ensure that brain pressure remains within safe limits \cite{czosnyka2004monitoring}.
Other modalities include respiratory monitoring, used in critical care or post-operative contexts \cite{nicolo2020importance}; 
oximetry, which provides real-time data on blood oxygen levels \cite{luks2020pulse}; 
and continuous blood pressure monitoring for sustained cardiovascular assessment \cite{armitage2023diagnosing}.
The continuous nature of these modalities offers clinicians a dynamic view of physiological parameters, enabling early intervention and the evaluation of therapeutic outcomes. 

Time-series data typically undergo preprocessing to remove noise and outliers and to address issues such as missing data, sparsity and irregular sampling \cite{che2018recurrent}. Following this, features are often extracted based on expert knowledge to ensure their relevance and accuracy \cite{zabihihyperensemble}. Beyond this expert-guided approach, several computational methods, including Fourier transformations, are employed. When coupled with machine learning techniques, these methods prove effective in feature extraction from time-series data \cite{walker2022dual}. Time dependencies can be managed during the feature extraction phase or a subsequent step, for instance, by employing LSTM to capture dependencies over time \cite{zheng2021predicting}.

Time-series data are often used for detecting early signs of patient deterioration or recovery, predicting outcomes and determining appropriate interventions and treatment plans. For example, the presence of rising inflammatory markers and a fever may prompt a change in antibiotic regimen for a patient being treated for infection \cite{watkins2010role}, while the resolution of previously abnormal blood tests may prompt a consideration of hospital discharge. 
Some studies have integrated audio data for various disease prediction tasks \cite{walker2022dual, ceccarelli2022multimodal, salekin2021multimodal}, while others have used vital signs, such as heart rate recordings and blood pressure measurements, in their models \cite{el2020multimodal, duvieusart2022multimodal, Grant2021}.

\subsection{Tabular data}
Tabular data refers to structured data that can be represented in a table format with rows and columns, generally containing one data point per feature per subject. This differs from time-series data which, although often stored in tabular format, capture continuous information over time for each subject. Tabular data offer a snapshot-like overview while time-series data provide a dynamic view of physiological or clinical variables. In healthcare, the systematic organisation and analysis of various forms of patient information often uses tabular data. For example, demographic data, such as a patient's age, sex and ethnicity, is often provided in tabular format and can provide essential context for individualised care \cite{duvieusart2022multimodal, xu2020accurately}. Health scoring systems, including standardised assessments like disease severity scores (e.g. APACHE II \cite{knaus1985apache}), pain scales (e.g. KOOS \cite{pierson2021algorithmic}) and quality-of-life indices (e.g. EQ-5D \cite{herdman2011development}) can offer quantifiable insights into a patient's condition and treatment efficacy. Clinical laboratory data such as blood chemistry analysis are another very common example and can be used to add granularity to assessments of patient condition and inform prognosis \cite{xu2020accurately}. Pharmaceutical data, such as drug dosages, formulations and pharmacological properties, are often found in tabular format and are vital for medication management and research \cite{Samak2020}.
Beyond these core elements, additional layers of tabular data can include insurance and billing information, care plans, social and behavioural data, medical history, medical inventory, clinical outcomes and follow-up data \cite{Joh20_MIV}. Other data modalities such as time-series data can often be aggregated and presented in tabular form for summary analysis \cite{duvieusart2022multimodal}. 

Tabular data provide a foundation for predictive models and decision support systems. These data are predominantly analysed using supervised techniques, such as linear and logistic regression \cite{polsterl2021combining}, decision trees and ensembles of them, like XGBoost and random forests \cite{krones2023physionet}, as well as support vector machines \cite{Nie2019}. Similar to time-series data, feature engineering emerges as an essential aspect for tabular data. Methods for feature selection such as correlation-based selection and feature extraction techniques such as principal component analysis serve to diminish data dimensionality, enhancing model performance \cite{Ayesha2021}.
Deep learning approaches have also made inroads into tabular data analysis, with models like TabNet \cite{arik2022tabnet} and 1D-CNNs \cite{ziv2022tabular}.

Tabular data play a significant role in a broad range of medical tasks, including but not limited to, disease prediction \cite{polsterl2021combining}, patient risk evaluation \cite{bagheri2020multimodal} and outcome forecasting \cite{vanguri2022multimodal}. 
For instance, tabular data was used with MRI for the detection of prostate cancer \cite{Reda2018}; with demographic and audio data for heart murmur detection \cite{walker2022dual};  with demographics, images and time-series data for Alzheimer's diagnosis \cite{el2020multimodal}; and with genomic data as well as  CT and pathological images for predicting lung cancer response to immunotherapy \cite{vanguri2022multimodal}.

\section{Model development}\label{sec:pipeline} 
Developing a multimodal deep learning framework in healthcare follows the same steps as most other machine learning approaches and consists of problem formulation, data pre-processing, model training and model evaluation \cite{crisp}. What makes the deep learning process unique is the notion of pre-training and fine-tuning \cite{Goodfellow-et-al-2016}, which separates the training procedure into two stages. Pre-training a model aims for a model to understand salient concepts in data, which is usually task-agnostic (Section\,\,\ref{sec:pretrain}). Fine-tuning on the other hand transforms these concepts into meaningful information that can be interpreted by humans (Section\,\,\ref{sec:finetuning}).

This section will focus on those two training stages--pre-training and fine-tuning--of the deep neural network model, transforming it from an initial random state to a learned state capable of executing specific tasks in downstream applications. In addition to the training stages, we will discuss the data pre-processing step, which is a prerequisite to be carried out whenever training is performed, and the model evaluation, which is performed after training but before deployment to the real-world.

\begin{figure}[t]
    \centering
    \includegraphics[width=\textwidth]{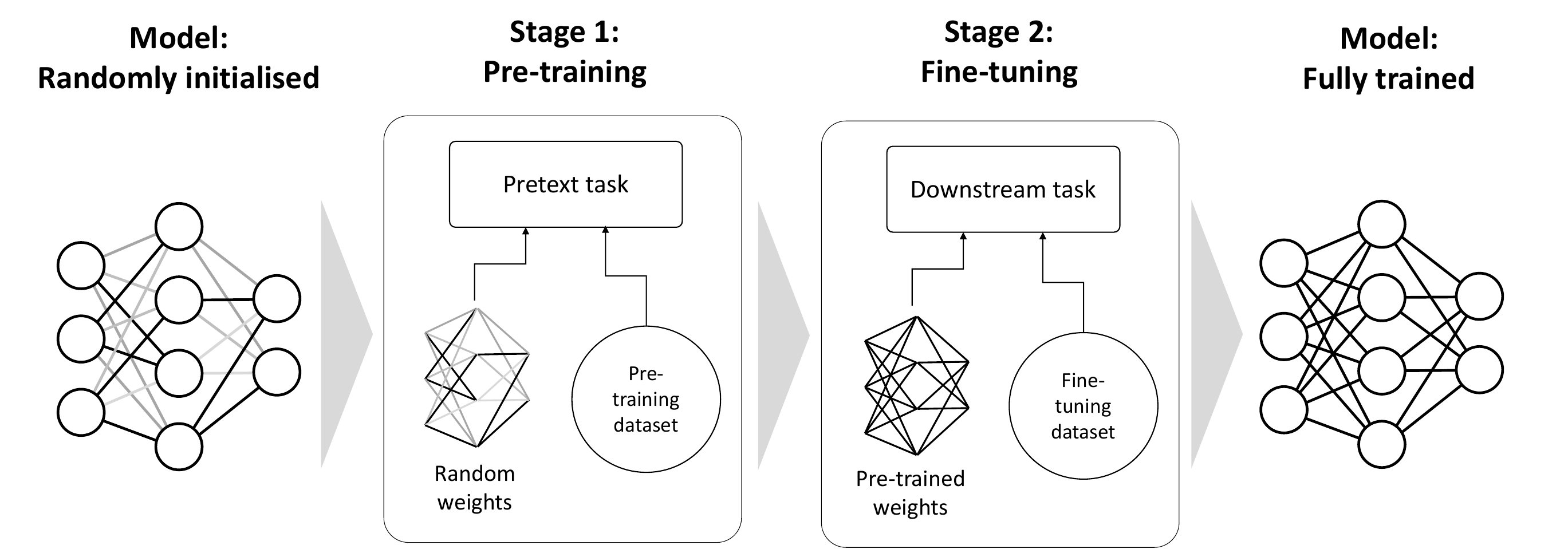}
    \caption{\textbf{Model development.} After pre-training the model, the model weights are fine-tuned on the target domain (e.g. medical images) and the model architecture is adjusted to the target task (e.g. classification).}
    \label{fig:pipeline_example}
\end{figure}

\subsection{Data pre-processing}
{\it Data pre-processing} serves as a critical phase in machine learning pipelines, transforming raw and unstructured data into a format optimised for subsequent model training \cite{nawi2013effect, maharana2022review}. This transformation typically follows data collection and precedes the training of machine learning models.

The process encompasses data cleaning, integration, transformation and reduction. These steps aim to handle missing values, remove inconsistencies and convert categorical variables into numerical form \cite{little2019statistical, maharana2022review}. By enhancing data quality, pre-processing facilitates better pattern recognition and relationships within the data, ultimately contributing to improved model performance, reduced overfitting, and more reliable and interpretable results.

In {\it data cleaning}, the aim is to impute missing values and remove duplicates from the raw dataset \cite{impute}. In some cases, outliers are removed ensuring that these outliers do not contain information that represents a meaningful edge case. For example, a fully blacked-out image or a blank text file can be considered to be an outlier as they do not contain any useful information \cite{yu2002findout}. 

{\it Data integration} involves combining and merging data from multiple sources or datasets to create a unified and comprehensive dataset \cite{duvieusart2022multimodal}. This process aims to facilitate more comprehensive analysis, modeling and decision-making. For example, integrating multimodal data from multiple body parts to train a proposed framework can be especially useful for a model to be more robust towards out-of-distribution data (i.e. data that are fundamentally different from the data on which a machine learning model has been trained) \cite{chen2022pan}.

{\it Data transformation} involves altering the format, structure or values of data to make them more suitable for analysis or modeling. This may involve obtaining vector embeddings for raw text data \cite{devlin2018bert} or resizing images for image data \cite{AlexNet12}. This process aims to improve data quality and prepare the data for specific machine learning algorithms. It often involves data scaling (bringing numerical variables to a similar scale or range) and data normalisation to allow for more efficient training \cite{ioffe2015batch}. When considering inputs from multiple modalities, normalisation enables the model to treat each data stream fairly, preventing any single modality from exerting undue influence on the overall input.

Finally, {\it data reduction}, which consists of feature selection or dimensionality reduction, is carried out in cases where the data remain too noisy after cleaning and normalisation or where the dimensionality of the data points may be too high for the model. Dimensionality reduction can be performed using techniques like principle component analysis \cite{pca}. Data reduction steps convert the data into a trainable state and thereby enable neural networks to interpret the data in an optimal manner.

\subsection{Stage 1: Model pre-training}\label{sec:pretrain}
{\it Model pre-training} (or {\it representation learning}) aims to build strong modality-specific or joint representations of the input data by training a neural network to perform a specific {\it pretext task} (i.e. a task involving a loss function that allows the network to learn these representations) \cite{shurrab2022self}. The representations of the input data are obtained by training the model on large datasets, such as ImageNet \cite{deng2009imagenet} for image-based tasks, BookCorpus \cite{devlin2018bert} for text-based tasks or MIMIC \cite{johnson2023mimic} for medical images, and they provide a strong initialisation point for fine-tuning networks, typically on a much smaller dataset. 

Pre-training can itself contain multiple stages. For example, a model that has already been trained on natural images can be further pre-trained on a medical domain-specific dataset to provide an even stronger initialisation point for later fine-tuning on another dataset in the medical domain \cite{kalapos2022self}. This intermediate step is sometimes interpreted as a fine-tuning step. However, in this review we define fine-tuning as updating the model only for the required downstream task and pre-training as the stage that provides the initialisation point for fine-tuning the model. 

The pre-training learning strategy can also be further categorised into supervised and unsupervised (or  self-supervised) learning.

\noindent
\textbf{Supervised pre-training.} In {\it supervised learning} a model is trained on a large dataset that contains labels and uses the associated label as a supervisory signal \cite{hastie2009overview}. The pretext task in supervised learning is a simple classification or regression task based on the label. A common example is the training of deep networks on ImageNet-1k \cite{deng2009imagenet} using the given class labels per data-point to perform a classification task. As the dataset is large, the model will be able to generalise a given class based on the multiple examples the model sees under that class. In a multimodal setting, a signal from a single modality may act as a supervisory signal for another modality from the same data point \cite{radford2021learning, chen2023vlp}.

Supervised pre-training has proven to be strong on very large datasets such as ImageNet-21k \cite{ridnik2021imagenet}, mainly due to the generalisation achieved through the exposure to many examples. However, models trained using self-supervised learning have shown to be able to outperform models employing supervised learning when trained on smaller datasets such as ImageNet-1k \cite{atito2021sit}. This is because the learning capacity of supervised learning is limited, as it is constrained by the information contained in the attached supervisory signals (labels). Moreover, supervised pre-training is also disadvantageous due to the labour costs incurred through the need to label large scale datasets \cite{krishnan2022self}.

\noindent
\textbf{Self-supervised pre-training.}
To mitigate the disadvantages of supervised pre-training, {\it self-supervised learning} aims to train neural networks to learn data representations by leveraging the inherent properties of the data \cite{ericsson2022self}. Hence, the data do not need to be associated with a specific label; consequently, the learned representations can be more generalisable \cite{hendrycks2019using, azizi2021big, Ericsson_2021_CVPR}.

There exists a large number of pretext tasks for self-supervised learning that do not involve pre-defined labels, with commonalities and differences \cite{krishnan2022self}. Here we have grouped them into two categories: discriminative and restorative. In {\it discriminative tasks}, the loss function of the pretext task guides the model to cluster similar data points of the dataset together \cite{haghighi2022dira}. One way this can be performed is by contrastive learning \cite{sowrirajan2021moco}, where specific loss functions enable association between similar observations and disassociation between dissimilar ones. 
{\it Restorative tasks} aim to learn representations of data by forcing the neural network to reconstruct a given raw data point based on some reduced or corrupted representation of that data point \cite{haghighi2022dira}. For example the language model GPT-3 \cite{brown2020language} was trained to predict the next word in partially masked sentences, while image-based model have been trained to reconstruct an image when provided with a distorted one \cite{atito2022gmml}.

\subsection{Stage 2: Model fine-tuning}\label{sec:finetuning}
Once pre-training has been performed, the trained model is expected to hold rich information about the semantic concepts in the pre-training dataset \cite{hendrycks2019using2}. This makes it possible to use the learned parameters (weights) from that model for multiple downstream tasks. The literature has shown that if the distribution of the downstream dataset aligns closely with the pre-training dataset, a model pre-trained on medical images is likely to outperform models pre-trained on natural images during fine-tuning, provided the datasets are of comparable size \cite{ma2022benchmarking}. Unlike pre-training, {\it fine-tuning} does not involve pretext tasks and instead is carried out for a specific downstream task that is formulated based on the problem at hand. This usually involves two things: adjusting the model to the target task structure (e.g. for a classification problem, adjusting the last layer of the model); and training the model on the target data, which can either mean training all weights or only weights of specific layers. For multimodal data, modality-specific pre-trained models can be combined and then fine-tuned for a downstream task that involves multimodal data inputs.

\subsection{Model evaluation}
{\it Model  evaluation} is carried out prior to model deployment to real-world scenarios. It aims to test the model under various circumstances to ensure its effectiveness, reliability and generalisability \cite{widner2023lessons, beede2020human}. 

Deployment may expose the model to a substantial amount of out-of-distribution data that does not align with the training data's distribution and so evaluating the model's performance across various patient subgroups, diverse conditions and different geographical regions is essential to test its reliability and generalisability \cite{azizi2022robust}. This becomes particularly critical in a multi-site setup where data can significantly vary across healthcare facilities and populations \cite{tran2022plex, widner2023lessons, beede2020human}. Randomised controlled trials would form an ideal testing scenario, however, these have only been conducted by a handful of studies and are challenging to conduct \cite{han2023randomized}.
In machine learning, the concerns mentioned above can be categorised as the following key measures: robustness to data noise, interpretability and generalisability.

\textit{Robustness to data noise} refers to a machine learning model's ability to maintain stable and accurate performance even when the input data contains irrelevant or erroneous information, often referred to as `noise' \cite{tran2022plex}. It is related to an understanding of how much noise affects predictions (i.e. the uncertainty \cite{abdar2021review}), which subsequently contributes to clinical decision-making \cite{chua2023tackling}. There are various approaches to evaluating uncertainty for deep neural networks including Bayesian and ensemble methods and test-time augmentations of data \cite{gawlikowski2023survey}.

\textit{Interpretability} refers to the degree to which a machine learning model's predictions and decisions can be understood and explained by humans \cite{reyes2020interpretability}. It involves making complex and often opaque machine learning models more transparent and comprehensible to users, domain experts and stakeholders. Interpretable models provide insights into how they arrive at specific outcomes, which variables or features are influential and why certain decisions are made. Interpretability is crucial in healthcare as it helps build trust and ensures that the model's behaviour aligns with ethical and legal standards \cite{voigt2017eu}. A common example in the medical domain is the use of saliency maps to indicate where a pathology lies within a given medical image \cite{haghighi2022dira}. More recently, the inherent attention mechanism that exists in transformers \cite{vaswani2017attention} has allowed for improved interpretability of deep transformer networks \cite{caron2021emerging}.

\textit{Generalisability} of a machine learning model refers to the ability of a trained model to perform effectively and accurately on new, unseen data or with patient populations that were not part of the model's training dataset. In the context of pre-training models, generalisability can be evaluated by performing zero-shot-predictions, where predictions are made on new datasets that the model has not seen before \cite{tran2022plex}. Achieving strong generalisability is vital in healthcare applications if models are to provide valuable support across clinical environments and patient demographics while maintaining high performance \cite{widner2023lessons}. Common methods for assessing the generalisability of machine learning models include evaluating them on a test set of unseen data \cite{azizi2022robust}, testing them on diverse datasets sourced from various institutions \cite{azizi2022robust} and conducting clinical validation and expert review \cite{widner2023lessons}.

\section{Fusion approaches}\label{sec:concepts}
Here we describe basic concepts behind data fusion and multimodal deep learning. 

\begin{figure}[t]
    \centering
    \includegraphics[width=\textwidth]{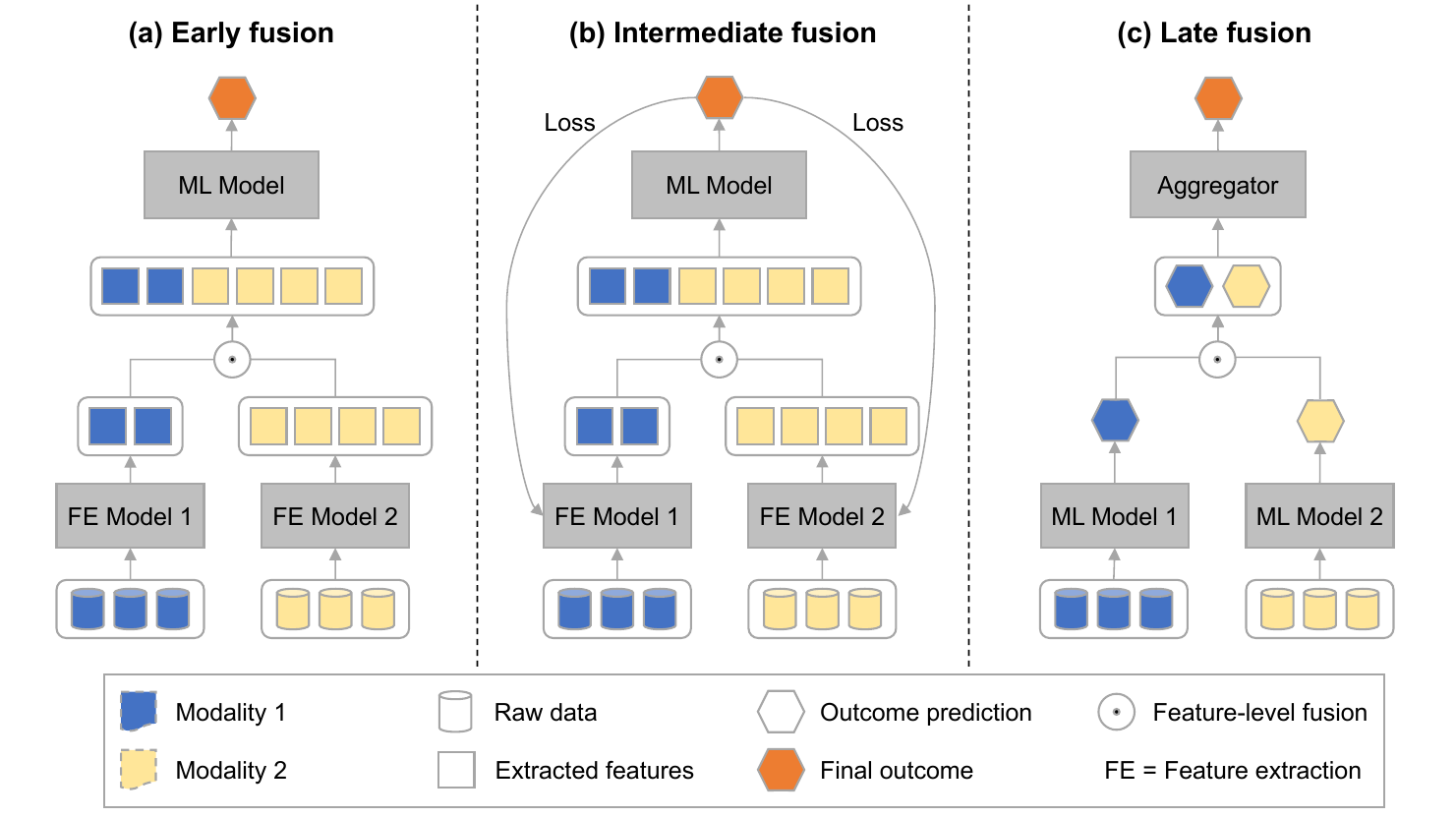}
    \caption{{\bf Data fusion architectures:} (a) Early fusion combines raw features or extracted features before passing them into the final model. The feature extraction is optional; (b) Intermediate fusion concatenates features extracted from the original data using an integrated modelling approach where the loss is back-propagated through the whole model; (c) In late fusion 
    the predictions or features are generated by multiple models and aggregated after their individual processing.}
    \label{fig:fusion}
\end{figure}

\begin{figure}[t]
    \centering
    \includegraphics[width=\textwidth]{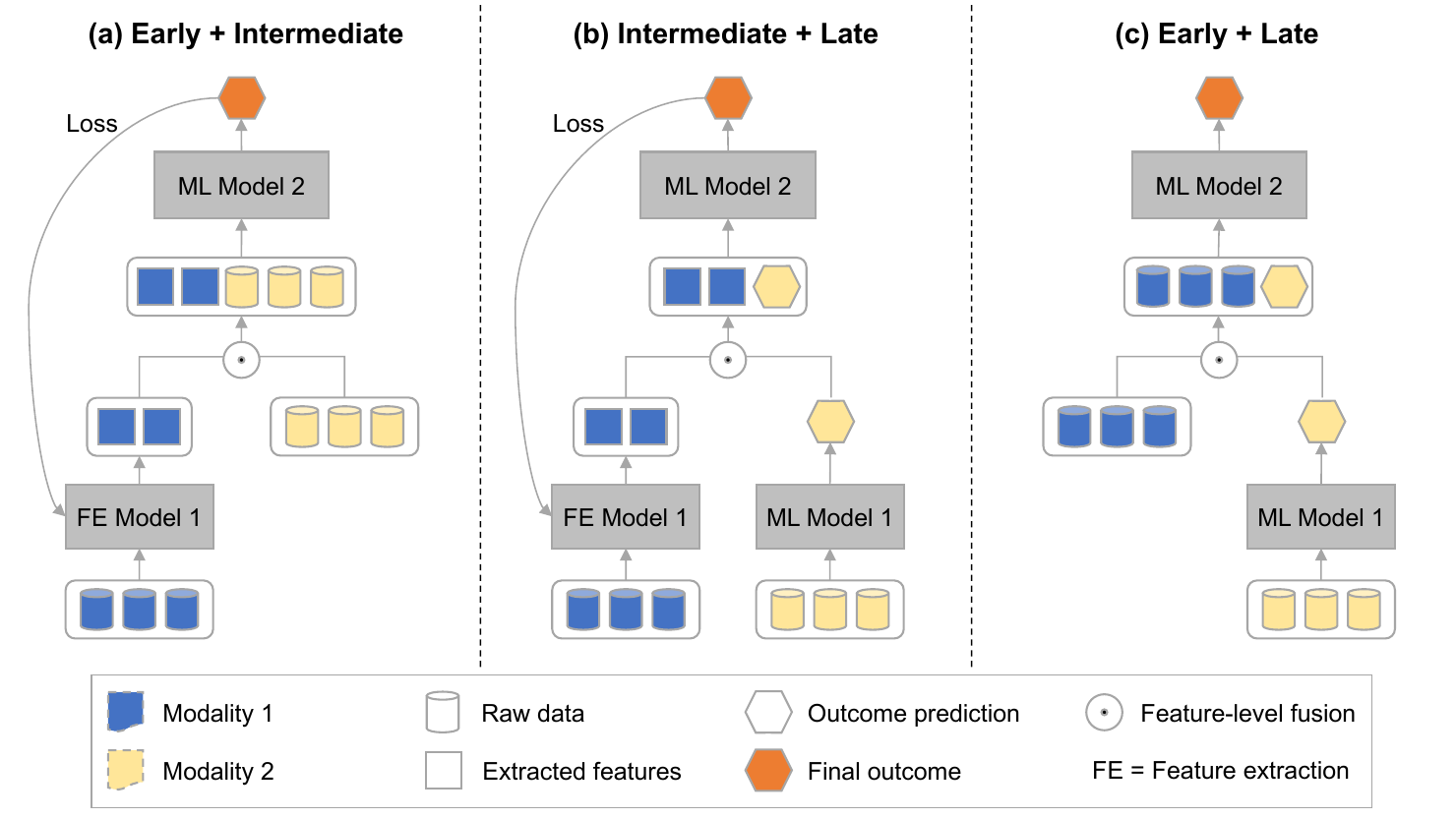}
    \caption{{\bf Examples of mixed fusion architectures:} (a) The loss is only back-propagated for some modalities (blue) while others (yellow) are fused at a later step; (b) Similar to (a), but predictions from only one modality are used; (c) Features from one modality (blue) are combined with predictions from another modality (yellow).}
    \label{fig:mixed_fusion}
\end{figure}

\subsection{Modality-level fusion}\label{sub:fusion}
Numerous taxonomies exist in the scientific literature for describing fusion approaches. Depending on how modalities are combined for a final prediction, the following techniques can be distinguished (\autoref{fig:fusion}): {early fusion}, {intermediate fusion}, {late fusion} and mixed approaches \cite{qiu2022multi}. 

\smallskip

\noindent
\textbf{Early fusion.} 
In {\it early fusion} \cite{Huang2020_2, huang2020fusion, kline2022multimodal} (also referred to as {\it data-level fusion} \cite{Ramachandram2017} or {\it feature-level fusion} \cite{Ayesha2021, dolly2019survey, hermessi2021multimodal}) multiple input modalities are combined before training a single machine learning model (see Figure\,\ref{fig:fusion}(a)). The data can be used in its raw form or may require various steps of feature extraction. This can range from simple aggregation methods to the use of separate models. 

The method of combining features also depends on the specific modalities and models involved. For instance, time-series data may need to be aggregated before use with models such as XGBoost \cite{walker2022dual}, while the combination of multiple images may involve stacking them as channels in a Convolutional Neural Network (CNN) framework \cite{taleb2021multimodal}.
One of the challenges in multimodal deep learning occurs when one attempts to combine data from multiple modalities at an early layer in the network, which can result in an imbalance of `data richness' from each modality \cite{li2022mplug}. For example, with vision and language inputs, both the vision and the language data are passed through light feature extractors to convert the data to the desired feature space. However, the vision features require more processing by the machine learning model compared to the language features, primarily due to the data richness present in the image itself. Consequently, the machine learning model would allocate disproportionate attention to the vision modality.
Moreover, the low-level features of a single modality may not necessarily provide any semantic meaning in a way that can be fused with another modalities. For instance, word embeddings (low-level features used for language data \cite{Vaswani2017}) lack the capability to deduce information from edges in an image, which are typically the kind of features extracted in computer vision models such as AlexNet \cite{AlexNet12}. This necessitates modality-specific feature encoding to standardise features in terms of semantic richness before forwarding them to a machine learning model for task-specific processing.

Early fusion has been implemented in clinical machine learning across multiple settings, such as for predictive tasks in cardiological \cite{walker2022dual, Brugnara2020}, oncological \cite{vanguri2022multimodal, Nie2019, silva2020pan} and neurological domains \cite{Li2019, Achalia2020}.
In \cite{duvieusart2022multimodal} an XGBoost model was used to integrate laboratory results, demographic information, vital signs and image features for cardiomegaly prediction. Similarly, \cite{Hyun2019} fused features from CT and PET scans with demographic data for lung cancer diagnosis and \cite{Dwivedi2022} fused MRI and PET images with demographic and genetic data for Alzheimer's disease prediction.

\smallskip
\textbf{Intermediate fusion.} 
In {\it intermediate fusion} \cite{kline2022multimodal, Ramachandram2017} (also referred to as {\it joint fusion} \cite{Huang2020_2, huang2020fusion}), different data modalities are first processed by individual models, before the extracted features are combined and fed into a final prediction model (see Figure\,\ref{fig:fusion}(b)). Unlike early fusion, here the loss function is back-propagated through the feature extraction model in order to generate improved feature representations during each iteration of training \cite{huang2020fusion}.
 
A common training approach for intermediate fusion involves first pre-training models separately on individual modalities \cite{duvieusart2022multimodal, Rajalingam2018, palepu2022tier}. Subsequently, the weights of these models are frozen, their outputs concatenated and the final model is trained. During the final training step, certain weights may remain fixed for a specific duration or within specific modalities. However, it is crucial for a process to be considered intermediate fusion that at least some weights are unfrozen at least once during the training process \cite{huang2020fusion}.
Intermediate fusion has its own complexities. The additional interactions and combinations can make the model complex and data-hungry \cite{huang2020fusion}. Deep learning models, which are often used in intermediate fusion, typically require a large amount of data to effectively learn from the intricate feature interactions.

Intermediate fusion has been used in cancer prediction, where it is common practice to combine pathological images with additional demographic and genomics data \cite{li2020novel, cheerla2019deep, schulz2021multimodal}. 
Similarly, within cardiology intermediate fusion has been used to combine X-ray images with demographic information, biomarkers and clinical measurements \cite{chauhan2020joint, duvieusart2022multimodal, Baltruschat2019}.
In addition, intermediate fusion has been used to predict brain disorders by combining MRI images with demographic information and other clinical and genetic information \cite{ghosal2021g, polsterl2021combining, Spasov2018, el2020multimodal}.

\medskip

\textbf{Late fusion.} 
In {\it late fusion} \cite{Huang2020_2, huang2020fusion, kline2022multimodal} (also referred to as {\it decision level fusion} \cite{Ayesha2021, behrad2022overview, dolly2019survey}), distinct models are run on separate modalities and the resulting predictions are merged through an aggregation function or an auxiliary model \cite{Baltrusaitis2018, Gao2020} (see Figure\,\ref{fig:fusion}(c)).

One advantage of late fusion is that it can easily deal with missing data for patients. For example, CLIP (Contrastive Language-Image Pre-Training) a network pre-trained using image and text data, is able to perform as an image classifier even without existing text data being provided as an input during inference (zero-shot learning) \cite{radford2021learning}. 
On the other hand, with late fusion it is not possible to model interactions and relationships between different modalities, which may lead to a loss of information \cite{Gao2020, Huang2019, huang2020fusion}.
Further, late fusion has its own set of challenges. The integration of results from different models can be complex and determining the optimal way of combining them is not always straightforward \cite{salekin2021multimodal}.

 Late fusion has been used in the diagnosis of cognitive impairment by combining MRI image predictions with demographic data and cognitive assessment scores \cite{Qiu2018}; in cancer prediction by merging MRI images with various biomarkers \cite{Reda2018}, and in COVID prediction by integrating CT scans with demographic information and clinical measurements \cite{zhou2021cohesive}.

\medskip

\textbf{Mixed fusion.} The previously discussed fusion methods can be combined (or mixed) in a way such that each modality is processed in its most optimal way, mitigating the modality imbalances caused by early fusion, but also modelling the inter-modality dependencies which cannot be done using late fusion. For example, when processing vision-language models, it has been found that modelling data jointly works best when image data has been independently processed to a certain degree before combining with language data \cite{li2022mplug, xu2023mplug}.

Whilst mixed fusion comes with the advantage of handling modality imbalances (as illustrated in Figure \ref{fig:mixed_fusion}, which allows for tailored integration of data modalities), it is challenging to design these networks as deciding at which point into the processing pipeline the modalities need to be combined requires careful consideration \cite{xu2023bridgetower}. 

In healthcare, mixed fusion mostly occurs when one of the data modalities is an image. This is common for multi-modal histomics data processing, where histopathology images are processed up to the global feature level before combining with tabular genomic data \cite{chen2022pan, Chen_2021_ICCV}. Other studies used various fusion techniques for a specific disease diagnosis, combining features from MRI images with demographic and genetic data and clinical measurements \cite{venugopalan2021multimodal}.

\subsection{Feature-level fusion}
\begin{figure}[t]
    \centering
    \includegraphics[width=\textwidth]{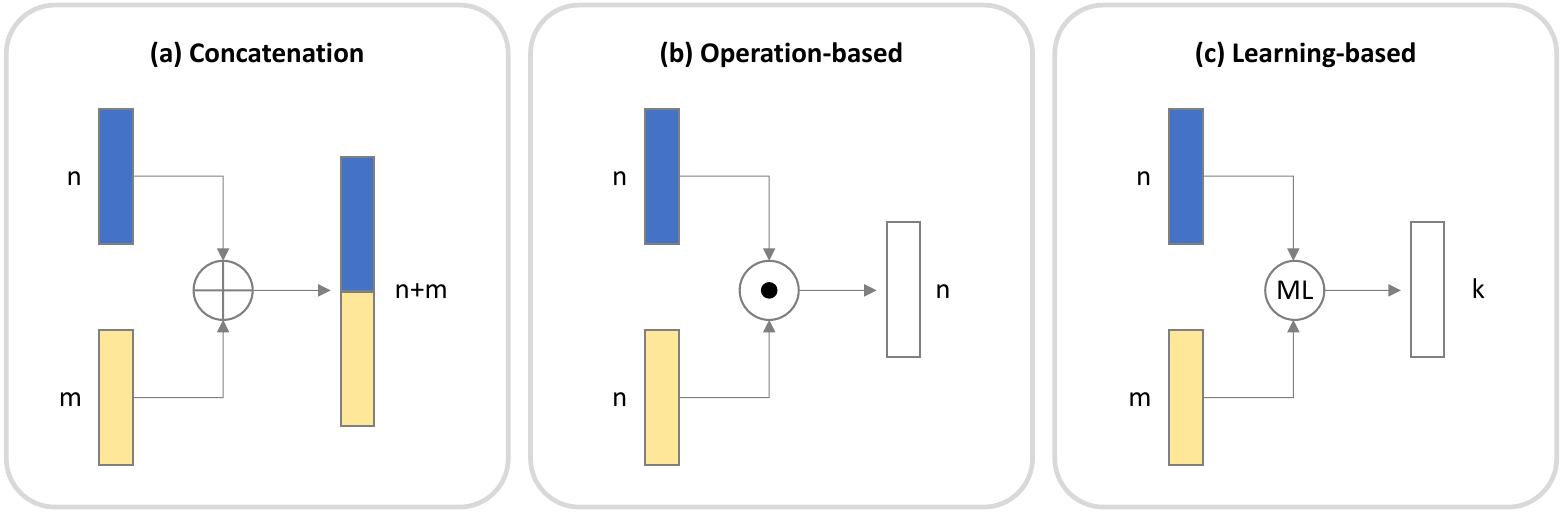}
    \caption{\textbf{Feature level fusion:} (a) Concatenation involves merging feature vectors end-to-end. (b) Operation-based methods combine vectors via element-wise mathematical operations or attention mechanisms, necessitating same-shaped vectors. (c) Learning-based fusion uses machine learning to reconstruct original features in a shared informative space.}
    \label{fig:feature_fusion}
\end{figure}

For early and intermediate fusion, various methods for fusing features can be distinguished. While we only provide a brief summary (\autoref{fig:feature_fusion}), a more extended overview can be found in the literature \cite{cui2022deep, Gao2020, Baltrusaitis2018, Ramachandram2017}.

\medskip
\textbf{Concatenation.} 
{\it Concatenation} is carried out when feature vectors are appended to form a single, longer vector \cite{cui2022deep} (see Figure\,\,\ref{fig:feature_fusion}(a)).

Since this approach is straightforward, flexible to different lengths of input and does not require additional parameters to tune, it is frequently used in practice \cite{holste2021end}. However, it has its drawbacks: concatenating features creates long vectors, which may lead to overfitting, especially when training datasets are not sufficiently large \cite{duvieusart2022multimodal, walker2022dual, yan2021richer}.

Concatenation was used in  \cite{duvieusart2022multimodal} to combine biomarkers extracted from images with demographic information, aggregated vital signs and lab results. Similarly,  \cite{el2020multimodal} combined various features extracted from time-series data in Alzheimer's disease progression detection.

\medskip

\textbf{Operation-based.} In {\it operation-based} fusion, feature vectors are combined by element-wise operations, i.e. operations are performed on corresponding elements in two or more arrays or matrices of the same dimensions (see Figure\,\,\ref{fig:feature_fusion}(b)). 

This approach requires vectors to have the same shape, using either element-wise or channel-wise multiplication. Element-wise multiplication in attention layers refers to multiplying corresponding elements of two matrices, whereas channel-wise multiplication involves multiplying entire channels (a specific dimension in a multi-dimensional array) of one matrix with another, treating the channel as a single entity. Alternatively, feature vectors can be combined in an attention-based manner (using attention layers) \cite{cui2022deep, silva2020pan, schulz2021multimodal, polsterl2021combining}, using one feature vector as attention weights for the other. Moreover, in tensor-based multiplication, feature vectors are combined by conducting outer products, with the aim of providing more information beyond that of the individual features \cite{chen2020pathomic, cui2022deep, braman2021deep}.

Examples include \cite{chen2020pathomic} where the authors used tensor-based fusion to correlate pathological images with genomics data to improve diagnostic accuracy; and \cite{schulz2021multimodal} where attention layers were used for cancer prognosis prediction.

\medskip

\textbf{Learning-based.} 
In {\it learning-based} fusion, feature vectors are combined using a machine learning model (see Figure\,\ref{fig:feature_fusion}(c)). 

A specific example of learning-based methods is the Graph Convolutional Network (GCN), which employs nodes and edges to establish relationships between input data (e.g. nodes representing image features and edges signifying similarity between image and non-image features), ultimately learning a common feature vector \cite{parisot2018disease, cao2021using, cui2022deep}. Similarly, in image-based networks (e.g. CNNs, ViTs), different `image channels' can be used to combine information from different modalities \cite{taleb2021multimodal}. 

In \cite{polsterl2021combining} the authors used an extension to conventional CNNs to combine 3D images with tabular data in Alzheimer's disease prediction; and \cite{parisot2018disease} used GCNs to combine MRI and clinical features in Alzheimer's disease prediction.

\section{Multimodal applications}\label{sec:application}

\subsection{Multimodal datasets}\label{sec:data}
In \autoref{tab:datasets} we provide an overview of various open-source, multimodal healthcare datasets and archives of datasets that contain multimodal information. We have primarily focused on multimodal content, which includes imaging data. These datasets contain demographic information and a subset contain extensive electronic health records comprising treatment histories, clinical metrics and in rare instances, full-text medical reports and vital sign time-series data.

\subsection{Multimodal studies}
We have conducted an overview of recent studies that employ multimodal approaches in healthcare. We categorise these studies into four groups: those predicting brain disorders (Table \ref{tab:studies1}), those making cancer predictions (Table \ref{tab:studies2}), those making predictions related to chest (Table \ref{tab:studies3}) or skin conditions and other diseases (Table \ref{tab:studies4}). Although our list is not exhaustive, we have focused on offering a comprehensive overview of noteworthy examples of multimodal deep learning applications in healthcare, organised by their respective application areas.

Most of the studies combined imaging data with tabular data (e.g. demographics, clinical measurements). The most commonly employed imaging modalities were MRI, followed by PET, CT and general X-ray imaging. Dermoscopic images were the predominant choice for skin disease studies. In some instances, studies incorporated additional assessment scores or markers extracted from images as tabular data alongside patient metadata. Only a limited number of studies integrated free text data from reports or time-series data, such as audio recordings or vital signs. Typically, these studies focused on feature extraction from time-series data rather than using the raw time-series. It is worth noting that many studies related to cancer and brain disorders used tabular features extracted from genomic data, evident in research areas including Alzheimer's diagnosis \cite{Chung2020, venugopalan2021multimodal} and breast cancer diagnosis \cite{liu2019association, duanmu2020prediction}.

The predominant analysis method used on imaging data were CNNs. For time-series data, simple feature extraction using aggregation methods, such as calculating the mean and standard deviation, were commonly applied. For analysing textual data in reports, studies typically used RNNs. There were various techniques used for tabular data, depending on the fusion technique applied.
Early and intermediate fusion were the most frequently used approaches. Late fusion was less common and typically involved independently calculated prediction scores for each modality before they were combined. The stage at which features were combined varied significantly across studies. Transfer learning has become a common strategy, particularly in studies that encounter limited data \cite{walker2022dual}. While the majority of studies employed supervised learning, interest in unsupervised and semi-supervised approaches due to the challenges of acquiring labelled data was noted \cite{taleb2021multimodal, parisot2018disease, Chung2020, chauhan2020joint}.

\section{Future research and conclusions}\label{sec:conclusion}
Having reviewed the relevant data modalities, model development strategies and fusion approaches in healthcare, we now discuss the potential directions for future research in the field.

\subsection{Multimodal data availability and integration} 
One of the major challenges in developing multimodal machine learning approaches for healthcare lies in the limited availability of multimodal datasets, comprising of diverse clinical modalities including clinical notes, images and time-series data, available for the same individual \cite{shaik2023survey}. However, acquiring multimodal datasets for machine learning in healthcare poses significant challenges related to privacy regulations, ethical and regulatory complexities, limited collaborations and resource constraints \cite{greenhalgh2017beyond, rajpurkar2022ai}.  From the technical point of view the difficulties in aligning different modalities (often linked with longitudinal nature of data) complicates even further the creation of multimodal datasets suitable for the use with machine learning. To overcome those obstacles would require the creation of clearer regulatory framework and more efficient collaboration between the healthcare and research communities.

\subsection{Foundation models and generalist medical AI} 
Foundation models, large, multipurpose models, typically pre-trained in a self-supervised manner on extensive datasets \cite{bommasani2021opportunities}, can potentially reshape medical AI \cite{moor2023foundation}. They have demonstrated remarkable versatility in various domains including natural language processing \cite{devlin2018bert, openaigpt} and multimodal tasks \cite{wang2022omnivl}. However, their adoption in the medical domain is still in its early stages. This is partly due to the challenges of accessing diverse medical datasets and the complexity of the healthcare realm. Most of the current medical machine learning models were developed to address very specific tasks, limiting their adaptability and flexibility. Recent advances in foundation model research, such as multimodal architectures \cite{acosta2022multimodal} and self-supervised learning techniques \cite{krishnan2022self}, could potentially address these challenges.

These developments may pave the way for generalist medical AI models, designed to handle a broad range of medical tasks and data types, including images, text and structured information \cite{tu2023towards, moor2023foundation}. 
Inspired by foundational models, generalist medical AI models aim to address healthcare challenges by offering adaptable and versatile solutions. These models have the potential to dynamically adapt to new tasks without extensive retraining \cite{brown2020language, alayrac2020self}, handle diverse data modalities and represent medical knowledge for advanced reasoning \cite{acosta2022multimodal}. However, realising this potential requires addressing substantial challenges, such as refining model interpretability and ensuring ethical and responsible AI deployment. By overcoming these obstacles, future research could unlock the promise of generalist medical AI models across various medical applications, potentially improving patient care and healthcare practices \cite{zhou2023foundation}.

\subsection{Deployment and clinical impact}
Deploying multimodal AI in healthcare faces key challenges, particularly in integrating AI into diverse healthcare workflows \cite{kelly2019key, rajpurkar2022ai}. Differences in patient management systems, clinical practices and digital infrastructures require adaptable and compatible systems. Regulatory hurdles are also amplified for multimodal AI due to the varied nature of data sources and the need for stringent privacy measures for each data type. Building trust in multimodal AI systems requires clear demonstration of their ability to handle diverse data securely and effectively. Addressing these specific challenges is critical for the successful deployment and widespread acceptance of multimodal AI in healthcare \cite{Handley_2023, nasss}.

\subsection{Conclusions}
In our comprehensive review of multimodal deep learning fusion approaches within healthcare, we identified several key insights. Modality-level fusion techniques, which include early, intermediate and late fusion stages, have been shown potential in improving predictive accuracy across many healthcare areas, especially in disease diagnosis and prognosis. 
Our study of multimodal datasets revealed important resources that bring together imaging data, demographic details, electronic health records and clinical measurements. It is noteworthy how disease prediction research has effectively used multimodal deep learning to achieve better results in diagnosis, treatment planning and survival rate predictions. The frequent use of MRI among imaging methods highlights its importance in the expansion of multimodal healthcare applications. Transfer learning has been crucial in overcoming challenges related to data scarcity. While supervised learning remains the primary approach, the popularity of semi-supervised and unsupervised methods is increasing. This reflects a growing interest in addressing healthcare issues where expert-annotated data are limited.


\begin{small}
\bibliographystyle{plain}
\bibliography{references} 
\end{small}

\begin{table}[t!]
    \centering
    \tiny
    \caption{\textbf{Multimodal healthcare datasets.} We identified more than one dozen publicly available dataset that include multimodal healthcare data.}\label{tab:datasets}
    \begin{tabularx}{\linewidth}{
        >{\hsize=0.8\hsize}X
        >{\hsize=1.2\hsize}X
        >{\hsize=1.2\hsize}X
        >{\hsize=1.2\hsize}X
        >{\hsize=0.8\hsize}X
        >{\hsize=0.8\hsize}X
        >{\hsize=0.8\hsize}X
        >{\hsize=0.8\hsize}X
        >{\hsize=1.2\hsize}X
        >{\hsize=1.2\hsize}X
    }
        \toprule
        {\bf Dataset name} & {\bf Use case} & {\bf Origin} & {\bf No. observations} & {\bf Time period} & {\bf Image data} & {\bf Text data} & {\bf Timeseries data} & {\bf Tabular data} & {\bf Link and source} \\
        \midrule
        MOST & Bones -- Osteoarthritis studies & University of Alabama at Birmingham, University of Iowa & > 4,000 patients & 2003 -- 2010 & MRI, X-ray & Reports &  & Demographics, risk factors, examination results & \url{https://most.ucsf.edu/} \cite{MOST2013} \\
        OAI & Bones -- Osteoarthritis studies & Various sites across North America & > 26M images from 4,796 patients & 2004 -- 2006 & MRI, X-ray &  &  & Demographics, clinical data, examination results & \url{https://nda.nih.gov/oai/} \cite{OAI2006} \\
        ABIDE I + II & Brain -- Autism studies & Various sites across North America & > 2,000 patients & na -- 2017 & MRI &  &  & Demographics, phenotypic info, intelligence scores, examination scores & \url{http://fcon_1000.projects.nitrc.org/indi/abide/} \cite{di2014autism} \\
        ADNI I - III & Brain -- Dementia detection & Various sites across North America & 4K MRI, 2K fMRI, 1.5K DTI, 2.5K PET images from 5K patients & 2004 -- 2022 & MRI, PET & Reports & Vital signs & Demographics, examination results, genetics & \url{https://adni.loni.usc.edu/} \cite{adni} \\
        OASIS-3  & Brain -- Dementia detection & Washington University Knight Alzheimer Disease Research Center & 2,842 MIR, 1,472 CT, 2,157 PET images from 1,379 patients & > 30 years before 2019 & MRI, CT, PET &  &  & Demographics, statistics from images, status info, longitudinal outcomes & \url{https://www.oasis-brains.org/} \cite{oasis} \\
        IDA (archive) & Brain -- Neuroscience disease prediction & Various, 143 studies & 85,007 patients in total & Various & MRI, CT, PET, EEG, SPECT &  &  & Demographics, genetics, exam scores & \url{https://ida.loni.usc.edu/login.jsp} \\
        Ischemic Stroke & Brain -- Stroke treatment outcome & University of Heidelberg & 246 patients & 2014 -- 2018 & CT &  &  & Demographics, clinical scores & \cite{Brugnara2020} \\
        BCDR & Breast -- Cancer analysis & Centro Hospitalar São João, at University of Porto & 1,734 patients & 1994 -- 2009 & Mammograms, Ultrasound images & Reports &  & Demographics, clinical history, segmentations & \url{https://bcdr.eu/} \\
        DDSM (archive) & Breast -- Cancer analysis & Various sites across North America & 2,620 studies & 1988 -- 1999 & Mammograms &  &  & Demographics, breast density rating, abnormally ratings & \url{http://www.eng.usf.edu/cvprg/Mammography/Database.html} \\
        TCIA (archive) & Cancer -- Archive of datasets & Various, 180 studies & 1 - 26,254 & Various & MRI, CT, PET, Pathology &  &  & Demographics, outcomes, genetics, treatment details & \url{https://www.cancerimagingarchive.net/} \\
        TCGA (archive) & Cancer -- Cancer analysis & Various sites across North America & 11,007 images from 86,513 patients & 2006 -- today & Histopathology & Reports &  & Demographics, omics, clinical data, examination results & \url{https://portal.gdc.cancer.gov/} \\
        CheXpert & Chest -- Image exploration & Stanford Hospital & 224,316 images from 65,240 patients & 2002 -- 2017 & X-ray &  &  & Demographics & \url{https://stanfordmlgroup.github.io/competitions/chexpert/} \cite{CheXpert19} \\
        MIMIC-IV & Chest -- Critical care data & Beth Israel Deaconess Medical Center emergency department & 69,619 unique ICU stays for 50,048 patients, 361,363 images from 64,586 patients & 2008 -- 2019, 2011 -- 2017 & X-ray & Reports & Vital signs, fluid inputs, etc. & Demographics, tracking data, EHR & \url{https://physionet.org/content/mimiciv/2.2/} \cite{Joh20_MIV, Joh19_CXR_Physio} \\
        NIH Chest X-ray & Chest -- Image exploration & NIH Clinical Center Bethesda, MD, USA & 108,948 images from 32,717 patients & 1992 -- 2015 & X-ray &  &  & Demographics & \url{https://www.kaggle.com/datasets/nih-chest-xrays/data} \cite{wang2017chestx} \\
        PADCHEST & Chest -- Image exploration & Hospital San Juan (Spain) & 160,868 images from 69,882 patients & 2009 -- 2017 & X-ray & Reports &  & Demographics & \url{https://bimcv.cipf.es/bimcv-projects/padchest/} \cite{bustos2020padchest} \\
        HAM10000 & Skin -- Lesions detection & Medical University of Vienna, Cliff Rosendahl in Queensland & 10,015 images & 1998 -- 2018 & Dermatoscopic images &  &  & Demographics & \url{https://dataverse.harvard.edu/dataset.xhtml?persistentId=doi:10.7910/DVN/DBW86T} \cite{HAM10000} \\
        Biobank (archive) & Various -- Disease prediction & United Kingdom & 500K patients & 2006 -- today & MRI, Ultrasound, X-ray &  &  & Demographics, genetics, exam scores, biomarkers, survey data, outcome data & \url{https://www.ukbiobank.ac.uk/} \\
        \bottomrule
    \end{tabularx}
\end{table}

\begin{table}[t!]
    \centering
    \tiny
    \caption{\textbf{Studies using multimodality approaches in brain disorder prediction.} We describe the fusion approach during fine-tuning, not pre-training; Early fusion can also involve heavy feature extraction using ML methods, the difference to intermediate fusion is that the loss is not backpropagated; While we only listed MRI as one modality, MRI can itself stand for various MRI modalities; `NN` as ML methods stands for various neural network architectures outside of the classical RNN, AE or CNN architectures; `Feature extraction` stands for various methods of feature extraction outside the mentioned architectures; GCN = Graph Convolutional Network; AE = Autoencoder; RF = Randon Forrest; GB = Gradient Boosting.}\label{tab:studies1}
    \begin{tabularx}{\linewidth}{
        >{\hsize=1.2\hsize}X
        >{\hsize=1.0\hsize}X
        >{\hsize=0.8\hsize}X
        >{\hsize=1.0\hsize}X
        >{\hsize=1.1\hsize}X
        >{\hsize=0.8\hsize}X
        >{\hsize=0.7\hsize}X
        >{\hsize=1.3\hsize}X
        >{\hsize=1.2\hsize}X
        >{\hsize=1.1\hsize}X
        >{\hsize=0.8\hsize}X
    }
        \toprule
        {\bf Topic} & {\bf Study, Year} & {\bf Training strategy} & {\bf Fusion technique} & {\bf Image modalities} & {\bf Timeseries modalities} & {\bf Text modalities} & {\bf Tabular modalities} & {\bf ML methods before fusion} & {\bf ML methods after/for fusion} & {\bf No. observations} \\
        \midrule
        Alzheimer  & \cite{parisot2018disease}, 2018 & Semi-supervised & Early & MRI &  &  & Demographics & Feature extraction & GCN & > 2,000 \\
        Alzheimer  & \cite{Chung2020}, 2020 & Semi-supervised & Intermediate & MRI &  &  & Demographics, genetic data & GCN & GCN & 1,675 \\
        Alzheimer  & \cite{Li2019}, 2019 & Supervised & Early & MRI & Cognitive measurements &  & Demographics, assessment scores & RNN, feature extraction & Cox regression & 822 \\
        Alzheimer  & \cite{Dwivedi2022}, 2022 & Supervised & Early & MRI, PET &  &  &  & Feature extraction, CNN & SVM & 200 \\
        Alzheimer  & \cite{zhou2019effective}, 2019 & Supervised & Intermediate & MRI, PET &  &  & Risk factors from genetic data & Feature extraction & NN & 805 \\
        Alzheimer  & \cite{Thung2017}, 2017 & Supervised & Intermediate & MRI, PET &  &  & Demographics, genetic data & Feature extraction & NN & 805 \\
        Alzheimer  & \cite{el2020multimodal}, 2020 & Supervised & Intermediate & MRI, PET & Vital signs &  & Demographics, assessment scores & Feature extraction, CNN, RNN, NN & NN & 1,536 \\
        Alzheimer  & \cite{Suk2014}, 2014 & Supervised & Intermediate & MRI, PET &  &  &  & DBM & SVM & 398 \\
        Alzheimer  & \cite{Spasov2018}, 2018 & Supervised & Intermediate & MRI &  &  & Demographics, genetic data, assessment scores & CNN, NN & NN & 376 \\
        Alzheimer  & \cite{polsterl2021combining}, 2021 & Supervised & Intermediate & MRI &  &  & Demographics, clinical measurements & CNN & NN & 1,341 + 755 \\
        Alzheimer  & \cite{venugopalan2021multimodal}, 2021 & Unsupervised, Supervised & Various & MRI &  &  & Demographics, clinical measurements, genetics & CNN, AE & NN & 2,004 \\
        Bipolar\hspace{1cm} disorder  & \cite{Achalia2020}, 2020 & Supervised & Early & MRI &  &  & Demographics, neurocognitive measures & Feature extraction & SVM & 60 \\
        Bipolar\hspace{1cm} disorder  & \cite{ceccarelli2022multimodal}, 2022 & Supervised & Late & Video & Audio recording & Transcripts &  & RNN & RF & > 150 \\
        Mild cognitive impairment & \cite{Qiu2018}, 2018 & Supervised & Late & MRI &  &  & Demographics, assessment scores & CNN, NN & Majority voting & 386 \\
        Multiple sclerosis conversion & \cite{Yoo2019}, 2019 & Unsupervised & Intermediate & MRI &  &  & Demographics, clinical measurements & Feature extraction, DBN, CNN & NN & 140 \\
        Schizophrenia  & \cite{ghosal2021g}, 2021 & Unsupervised & Intermediate & MRI &  &  & Genetic data & AE & NN & > 100 \\
        Coma outcome prediction & \cite{zheng2021predicting}, 2021 & Supervised & Late &  & EEG &  & Demographics & CNNs, RNNs & Majority voting & 1,038 \\
        \bottomrule
    \end{tabularx}
\end{table}

\begin{table}[t!]
    \centering
    \tiny
    \caption{\textbf{Studies using multimodality approaches in cancer prediction.} We describe the fusion approach during fine-tuning, not pre-training; Early fusion can also involve heavy feature extraction using ML methods, the difference to intermediate fusion is that the loss is not backpropagated; While we only listed MRI as one modality, MRI can itself stand for various MRI modalities; `NN` as ML methods stands for various neural network architectures outside of the classical RNN, AE or CNN architectures; `Feature extraction` stands for various methods of feature extraction outside the mentioned architectures; GCN = Graph Convolutional Network; AE = Autoencoder; RF = Randon Forrest; GB = Gradient Boosting.}\label{tab:studies2}
    \begin{tabularx}{\linewidth}{
        >{\hsize=1.2\hsize}X
        >{\hsize=1.0\hsize}X
        >{\hsize=0.8\hsize}X
        >{\hsize=1.0\hsize}X
        >{\hsize=1.1\hsize}X
        >{\hsize=0.8\hsize}X
        >{\hsize=0.7\hsize}X
        >{\hsize=1.3\hsize}X
        >{\hsize=1.2\hsize}X
        >{\hsize=1.1\hsize}X
        >{\hsize=0.8\hsize}X
    }
        \toprule
        {\bf Topic} & {\bf Study, Year} & {\bf Training strategy} & {\bf Fusion technique} & {\bf Image modalities} & {\bf Timeseries modalities} & {\bf Text modalities} & {\bf Tabular modalities} & {\bf ML methods before fusion} & {\bf ML methods after/for fusion} & {\bf No. observations} \\
        \midrule
        Brain tumor survival prediction & \cite{Nie2019}, 2019 & Supervised & Early & MRI &  &  & Demographics & CNN & SVM & 68 \\
        Brain tumor survival prediction & \cite{braman2021deep}, 2021 & Supervised & Intermediate & MRI, Pathological images &  &  & Demographics, clinical measurements, DNA results & CNN, NN & NN & 176 \\
        Breast cancer chemotherapy response prediction & \cite{duanmu2020prediction}, 2020 & Supervised & Intermediate & MRI &  &  & Demographics, genomic data & CNN, NN & NN & 112 \\
        Breast cancer & \cite{Yala2019}, 2019 & Supervised & Intermediate & X-ray &  &  & Demographics, risk factors & CNN & NN & 71M examinations of 89K patients \\
        Breast cancer & \cite{yan2021richer}, 2021 & Unsupervised & Intermediate & Pathological images &  &  & Structured data from EMR & AE, CNN & NN & 185 \\
        Breast cancer & \cite{liu2019association}, 2019 & Unsupervised & Intermediate &  & Genes &  & Demographics, genomic data & AE & AE & 24K genes for 1K patients \\
        Breast cancer survival prediction & \cite{li2020novel}, 2020 & Unsupervised & Intermediate & Pathological images &  &  & Genomic data & CNN, AE & AE & 826 \\
        Breast cancer & \cite{holste2021end}, 2021 & Supervised & Various & MRI &  &  & Demographics, clinical indications & CNN, NN & NN & 5,248 \\
        Carcinoma & \cite{Kharazmi2018}, 2018 & Unsupervised & Early & Dermatological images &  &  & Demographics & AE & Softmax classifier & 1,191 \\
        Lung cancer diagnosis & \cite{Hyun2019}, 2019 & Supervised & Early & PET, CT &  &  & Demographics & Feature extraction & Various & 396 \\
        Lung cancer response prediction to immuntherapy & \cite{vanguri2022multimodal}, 2022 & Supervised & Early & CT, Pathological images &  &  & Demographics, genomic data & Feature extraction & NN & 247 \\
        Pan-cancer prognosis & \cite{silva2020pan}, 2020 & Supervised & Early & Pathological images &  &  & Demographics, clinical measurements, DNA results & Feature extraction, CNN, NN & NN & 11,081 \\
        Pan-cancer prognosis & \cite{cheerla2019deep}, 2019 & Unsupervised & Intermediate & Pathological images &  &  & Demographics, genetics & NN, CNN & Siamese network & 11,160 \\
        Prostate cancer detection & \cite{rubinstein2019unsupervised}, 2019 & Unsupervised & Intermediate & PET, CT &  &  &  & AE & AE &  \\
        Prostate cancer detection & \cite{Reda2018}, 2018 & Unsupervised & Late & MRI &  &  & Biomarkers & Feature extraction, AE & AE & 18 \\
        Renal cancer prognosis prediction & \cite{schulz2021multimodal}, 2021 & Supervised & Intermediate & MRI, CT, Pathological images &  &  & Genomic data & CNN, NN & NN & 230 \\
        Tumor soft-tissue sarcoma segmentation & \cite{Guo2019}, 2019 & Supervised & Various & MRI, PET, CT &  &  &  & CNN & NN or RF & 50 \\
        Various cancer & \cite{chen2020pathomic}, 2020 & Unsupervised, Supervised & Early & Pathological images &  &  & DNA results & GCN, CNN, NN & Various & > 2,000 samples \\
        \bottomrule
    \end{tabularx}
\end{table}
\begin{table}[htpb]
    \centering
    \tiny
    \caption{\textbf{Studies using multimodality approaches in predicting chest related conditions.} We describe the fusion approach during fine-tuning, not pre-training; Early fusion can also involve heavy feature extraction using ML methods, the difference to intermediate fusion is that the loss is not backpropagated; While we only listed MRI as one modality, MRI can itself stand for various MRI modalities; NN as ML methods stands for various neural network architectures outside of the classical RNN, AE or CNN architectures; `Feature extraction` stands for various methods of feature extraction outside the mentioned architectures; GCN = Graph Convolutional Network; AE = Autoencoder; RF = Randon Forrest; GB = Gradient Boosting.}\label{tab:studies3}
    \begin{tabularx}{\linewidth}{
        >{\hsize=1.2\hsize}X
        >{\hsize=1.0\hsize}X
        >{\hsize=0.8\hsize}X
        >{\hsize=1.0\hsize}X
        >{\hsize=1.1\hsize}X
        >{\hsize=0.8\hsize}X
        >{\hsize=0.7\hsize}X
        >{\hsize=1.3\hsize}X
        >{\hsize=1.2\hsize}X
        >{\hsize=1.1\hsize}X
        >{\hsize=0.8\hsize}X
    }
        \toprule
        {\bf Topic} & {\bf Study, Year} & {\bf Training strategy} & {\bf Fusion technique} & {\bf Image modalities} & {\bf Timeseries modalities} & {\bf Text modalities} & {\bf Tabular modalities} & {\bf ML methods before fusion} & {\bf ML methods after/for fusion} & {\bf No. observations} \\
        \midrule
        Cardiomegaly & \cite{palepu2022tier}, 2022 & Self-supervised & Intermediate & X-ray &  & Reports &  & Transformer & Transformer & > 200,000 \\
        Cardiomegaly & \cite{duvieusart2022multimodal}, 2022 & Supervised & Intermediate and Early & X-ray & Vital signs & Reports & Demographics, clinical measurements & Feature extraction, CNN & NN & 2,774 \\
        Cardiovascular risk prediction & \cite{bagheri2020multimodal}, 2020 & Supervised & Early &  &  & Reports & Demographics, clinical measurements & RNN, FNN & NN & 5,603 \\
        Cardiomegaly & \cite{Grant2021}, 2021 & Supervised & Intermediate & X-ray & Vital signs &  & Demographics & Feature extraction, CNN & NN & 2,571 \\
        Cardiomegaly & \cite{Baltruschat2019}, 2019 & Supervised & Intermediate & X-ray &  &  & Demographics & Feature extraction, CNN & NN & 6,245 \\
        Endovascular treatment outcome & \cite{Brugnara2020}, 2020 & Supervised & Early & CT &  &  & Assessment scores & Feature extraction & GB & 246 \\
        Endovascular treatment outcome & \cite{Samak2020}, 2020 & Supervised & Intermediate & CT &  &  & Demographics, biomarkers & Feature extraction, CNN & NN & 500 \\
        Heart murmur detection & \cite{walker2022dual}, 2022 & Supervised & Early &  & Audio recording &  & Demographics & Bayesian ResNet & Bayesian ResNet, GB & 942 \\
        Pathway analysis & \cite{nishimori2021accessory}, 2021 & Supervised & Early & X-ray & ECG &  &  & CNN & CNN & 294 \\
        Pulmonary oedema assessment & \cite{chauhan2020joint}, 2020 & Semi-supervised & Intermediate & X-ray &  & Reports &  & CNN, Transformer & NN & 247,425 \\
        Covid detection & \cite{xu2020accurately}, 2020 & Supervised & Early & CT &  &  & Demographics, clinical measurements, lab results & CNN & Various & 689 \\
        Covid progression prediction & \cite{fang2021deep}, 2021 & Supervised & Early & CT &  &  & Demographics, clinical measurements & CNN, NN & RNN, NN & 1,040 \\
        Covid severity prediction & \cite{zhou2021cohesive}, 2021 & Supervised & Late & CT &  &  & Demographics, clinical measurements & CNN, NN & NN & 733 \\
        \bottomrule
    \end{tabularx}
\end{table}

\begin{table}[t!]
    \centering
    \tiny
    \caption{\textbf{Studies using multimodality approaches in predicting skin related conditions and other diseases.} We describe the fusion approach during fine-tuning, not pre-training; Early fusion can also involve heavy feature extraction using ML methods, the difference to intermediate fusion is that the loss is not backpropagated; While we only listed MRI as one modality, MRI can itself stand for various MRI modalities; NN as ML methods stands for various neural network architectures outside of the classical RNN, AE or CNN architectures; `Feature extraction` stands for various methods of feature extraction outside the mentioned architectures; GCN = Graph Convolutional Network; AE = Autoencoder; RF = Randon Forrest; GB = Gradient Boosting.}\label{tab:studies4}
    \begin{tabularx}{\linewidth}{
        >{\hsize=1.2\hsize}X
        >{\hsize=1.0\hsize}X
        >{\hsize=0.8\hsize}X
        >{\hsize=1.0\hsize}X
        >{\hsize=1.1\hsize}X
        >{\hsize=0.8\hsize}X
        >{\hsize=0.7\hsize}X
        >{\hsize=1.3\hsize}X
        >{\hsize=1.2\hsize}X
        >{\hsize=1.1\hsize}X
        >{\hsize=0.8\hsize}X
    }
        \toprule
        {\bf Topic} & {\bf Study, Year} & {\bf Training strategy} & {\bf Fusion technique} & {\bf Image modalities} & {\bf Timeseries modalities} & {\bf Text modalities} & {\bf Tabular modalities} & {\bf ML methods before fusion} & {\bf ML methods after/for fusion} & {\bf No. observations} \\
        \midrule
        Image modality combination & \cite{taleb2021multimodal}, 2021 & Self-supervised (via puzzle-solving) & Early & MRI, CT &  &  &  & CNN & CNN & > 350 \\
        Image recognition & \cite{Huang2021}, 2021 & Supervised & N/A & X-ray &  & Reports &  & CNN & Transformer & > 200,000 \\
        Microcytic hypochromic & \cite{Purwar2020}, 2020 & Supervised & Early & Pathological images &  &  & Blood test features & CNN & NN & 20 \\
        Mortality risk for ICU patients & \cite{Jin2018}, 2018 & Supervised & Intermediate &  & Vital signs & Reports &  & RNN, NN & NN & 42,276 \\
        Neonatal postoperative pain assessment & \cite{salekin2021multimodal}, 2021 & Supervised & Late & Video & Audio recording &  &  & CNN, RNN & Majority voting & > 200 \\
        Osteoarthritis & \cite{Tiulpin2019}, 2019 & Supervised & Early & X-ray &  & Reports & Demographics, assessment scores & Feature extraction, CNN & GB & > 5,000 \\
        Various diseases & \cite{Rod19}, 2019 & Supervised & Intermediate & X-ray &  & Reports & Demographics & Feature extraction, CNN & RNN & 8,530 \\
        Skin lesions classification & \cite{Yap2018}, 2018 & Supervised & Early & Dermatological images &  &  & Demographics & Feature extraction, CNN & NN & 2,917 \\
        Skin lesions classification & \cite{Gessert2020}, 2020 & Supervised & Intermediate & Dermatological images &  &  & Demographics & CNN, NNs & CNN ensemble & 25,331 \\
        Skin lesions classification & \cite{Kawahara2019}, 2019 & Supervised & Intermediate & Dermatological images &  &  & Demographics & CNN & NN & 1,011 \\
        \bottomrule
    \end{tabularx}
\end{table}
\end{document}